\documentclass[10pt,twocolumn,letterpaper]{article}

\usepackage{iccv}
\usepackage{times}
\usepackage{epsfig}
\usepackage{graphicx}
\usepackage{amsmath}
\usepackage{amssymb}

\usepackage{subfigure}
\usepackage{multirow}
\iccvfinalcopy 

\def\iccvPaperID{****} 

\begin{document}

\title{Learning to Predict More Accurate Text Instances for Scene Text Detection}
\author{XiaoQian Li\textsuperscript{\rm 1, \rm 2}\quad Jie Liu\textsuperscript{\rm 1}\quad ShuWu Zhang\textsuperscript{\rm 1, \rm 2}\quad  GuiXuan Zhang \textsuperscript{\rm 1}\\
\textsuperscript{\rm 1}{ Institute of Automation, Chinese Academy of Sciences,   Beijing 100190, China}\\
\textsuperscript{\rm 2} {School of Artiﬁcial Intelligence, University of Chinese Academy of Sciences,  Beijing 100049, China}\\
{\tt\small lixiaoqian2015@ia.ac.cn}
}

\maketitle
\begin{abstract}
 At present, multi-oriented text detection methods based on deep neural network have achieved promising performances on various benchmarks. Nevertheless, there are still some difficulties for arbitrary shape text detection,
especially for a simple and proper representation of arbitrary shape text instances. 
In this paper, a pixel-based text detector is proposed to facilitate the representation and prediction of text instances with arbitrary shapes in a simple manner.
Firstly, to alleviate the effect of the target vertex sorting and achieve the direct regression of arbitrary shape text instances, the starting-point-independent coordinates regression loss is proposed.
Furthermore, to predict more accurate text instances, the text instance accuracy loss is proposed as an assistant task to refine the predicted coordinates under the guidance of IoU.
%
 To evaluate the effectiveness of our detector, extensive experiments have been carried on public benchmarks which contain arbitrary shape text instances and multi-oriented text instances. 
 We obtain 84.8\% of F-measure on Total-Text benchmark.
 The results show that our method can reach state-of-the-art performance.

\end{abstract}

\section{Introduction}
Scene text detection, especially for multi-oriented text detection or arbitrary shape text detection, is one of the most challenging tasks in computer vision. Although the progress of deep learning has brought many technical improvements to scene text detection, there is still room for improvement in text detection due to the variability of texts, the complexity of background or the poor shooting circumstance.

The current mainstream approaches of text detection are based on generic object detection and image segmentation.
Recently, \cite{zhou2017east}, \cite{he2017deep}, \cite{zhong2018anchor}, \cite{zhang2019look} proposed pixel-based text detectors.
These pixel-based methods have solved the problem of complex manual anchor design of anchor-based methods, such as \cite{jiang2017r2cnn}, \cite{ma2018arbitrary}, \cite{liao2017textboxes}, \cite{liao2018textboxes++},  and obtained good performances. Besides, some segmentation-based methods combine with the geometric characteristics of text and have good performance in curved text.
Although the current approaches have obvious advantages in performance, there are still some deficiencies.

First of all, 
due to the difficulty in representing text instances of arbitrary shapes, some simple pixel-based models (such as \cite{zhou2017east}, \cite{he2017deep}) have deficiencies in curved text detection, and they cannot handle arbitrary shape text well, although they perform well in multi-oriented text detection.
Because the coordinates of concave polygons in these methods should be sorted in a specific order and the starting point of coordinate sequence should be determined, direct regression of curve text coordinates is limited.
Also, other methods, whether pixel-based methods such as \cite{zhang2019look} or segmentation methods such as \cite{long2018textsnake}, \cite{baek2019character}, \cite{zhu2018textmountain}, can deal with arbitrary shape text well, but the geometric properties of the text instances are required and increase the complexity and computation of the text detectors. Therefore, it is worth paying attention to the elegant representation of arbitrary shape text.



\begin{figure}
\centering
\includegraphics[width=8cm]{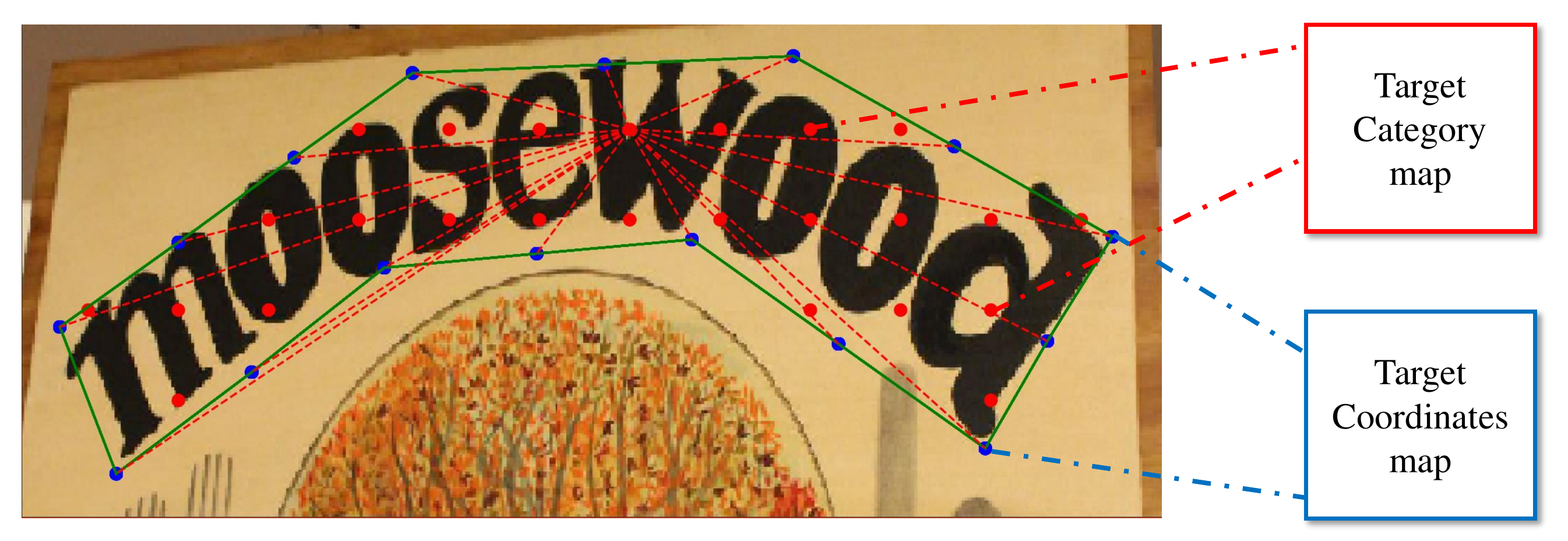} 
\hspace{0.00cm}
\caption{Label generation for pixel-based method. Red points are positive pixels, and blue points are vertices of polygon in green. The tasks of pixel-based method are to determine the category of pixels (corresponding to the target category map), and regress offsets between vertex and positive pixel (corresponding to the target coordinates map).  } \label{fig:lg}
\end{figure}

\begin{figure*}
\centering
\includegraphics[width=17.9cm,height=5cm]{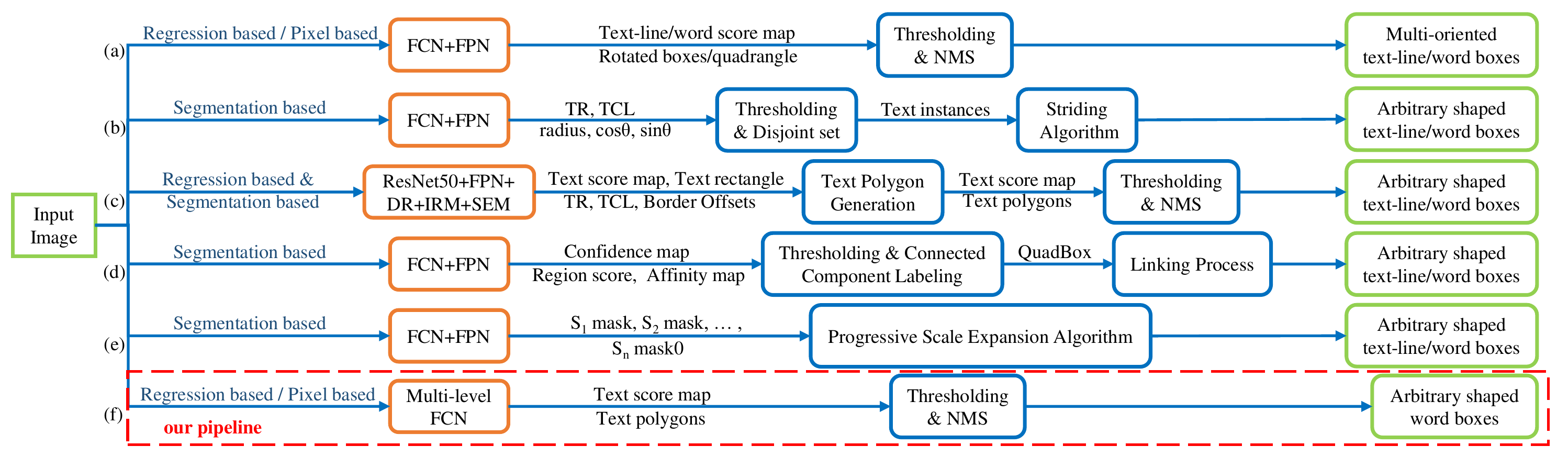} 
\hspace{0.00cm}
\caption{Comparison of pipelines in several recent works, TR means text region and TCL means text centerline. (a) The pipeline of pixel-based method \cite{zhou2017east} which is applicable to multi-oriented text detection; (b) the method in \cite{long2018textsnake} is based on image segmentation and feasible in arbitrary shape text detection; (c) the arbitrary shape text detector of LOMO \cite{zhang2019look} is based on segmentation (DR, IRM, SEM are short for direct regressor, iterative refinement module and shape expression module respectively); (d) the detector of \cite{baek2019character} predicts arbitrary shape text by predicting characters; (e) \cite{wang2019shape} proposed PSENet based on segmentation to detect curved text; (f) our pipeline is pixel-based and is simpler than previous detectors on arbitrary shape text.} \label{fig:pp}
\end{figure*}

Secondly, 
complex text representation increases the complexity of the method to a certain extent.
Some effective arbitrary shape text detectors are based on image segmentation such as \cite{zhang2019look}, \cite{long2018textsnake}, \cite{baek2019character}, \cite{wang2019shape}.
These methods require either additional post-processing step other than  Non-Maximum Suppression (NMS), or more complex post-processing steps to fuse the segmented components into text regions. Therefore, a simple and effective detector for arbitrary shape text is of much importance.

 Aiming at the above problems, a simple and effective pixel-based method  without additional post-precessing (except for NMS) in this paper is proposed. 
The model is implemented based on Single Shot Detector (SSD) \cite{liu2016ssd} with the backbone VGG16 \cite{simonyan2014very}, and achieves state-of-the-art performance. 

 Our main contributions can be summarized as follows:
\begin{itemize}

\item We propose the starting-point-independent regression loss instead of the conventional regression loss to optimize predicted coordinates of text instances, and different from segmentation-based methods, the coordinates of polygons can be directly optimized. 
\item The text instance accuracy loss is introduced to obtain text polygons with larger IoU, which further improves performance without increasing the computation of network.
\item A simple and effective pixel-based method is proposed, which only uses NMS post-processing step. The model is available for arbitrary shape text detection without additional annotation and obtains state-of-the-art performance on Total-Text dataset.
\end{itemize}

The remainder of this paper is organized as follows. In Section 2, we briefly introduce the related work on scene text detection. In Section 3, the proposed method will be demonstrated in details. In Section 4, the experimental results and analysis on several public datasets are presented. Finally, we summarize the proposed method.

\begin{figure*}
\centering
\includegraphics[width=1.9\columnwidth]{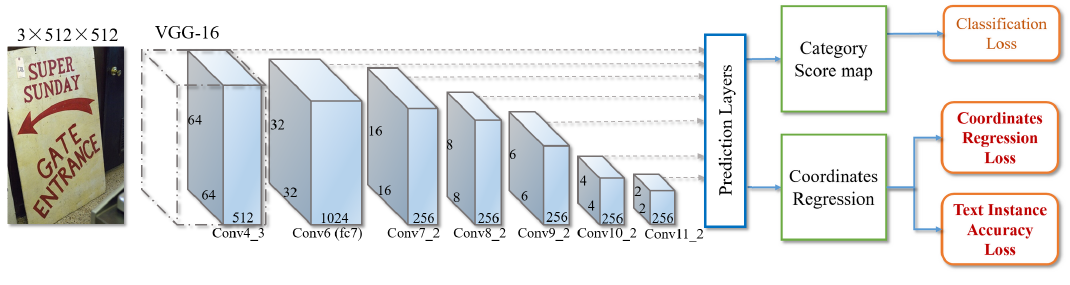} 
\hspace{0.00cm}
\caption{The proposed architecture. The outputs of the network are category score map and regressed coordinates, and the model is trained by classification loss, coordinates regression loss and text instance accuracy loss.} \label{fig:net}
\end{figure*}

\begin{figure}
\centering
\includegraphics[width=7.3cm, height =5.3cm]{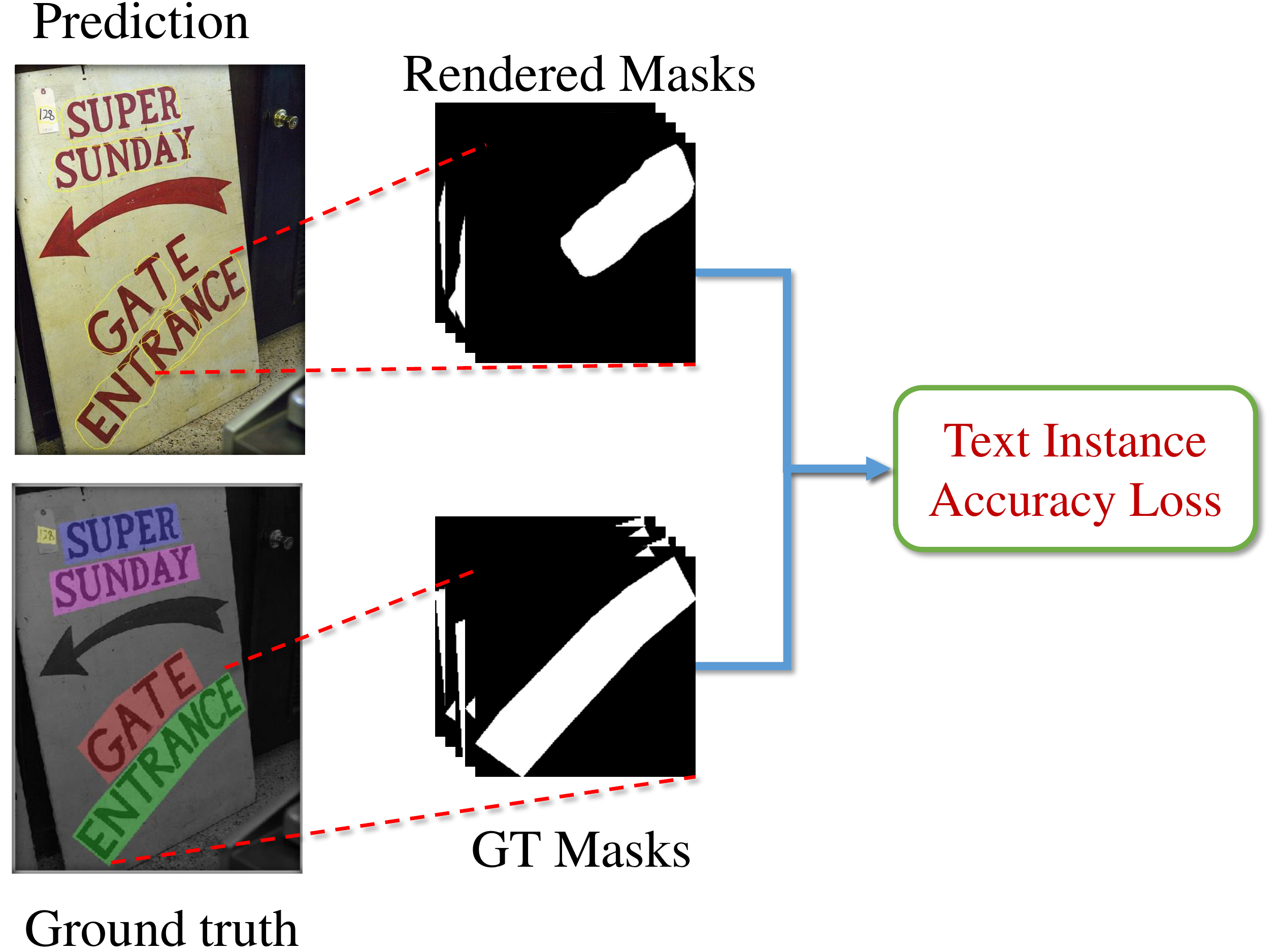} 
\hspace{0.00cm}
\caption{Text instance accuracy loss calculation. The predicted polygons are rendered into rendered masks and calculate the L1 loss between rendered masks and ground truth masks.} \label{fig:diff}
\end{figure}

\section{Related Work}
According to the type of text shape, this section introduces the related work on multi-oriented text detection and arbitrary shape text detection.

\textbf{Multi-oriented Text Detection.}
The development of generic object detection has led to the improvement of text detection. Taking text as a specific object,  \cite{jiang2017r2cnn}, \cite{ma2018arbitrary}, \cite{liao2017textboxes}, \cite{liao2018textboxes++} proposed text detectors based on generic object detection. According to characteristics of text instances, these methods designed a more appropriate anchor schema and modified convolutional kernel for text detection. Although anchors are well designed, there may still be missing matched text instances with extreme size. What's more, anchors should change as data changes, and more complicated anchor design may lead to more computation.
To eliminate the influence of anchor design, \cite{zhou2017east}, \cite{he2017deep}, \cite{zhong2018anchor} put forward pixel-based text detectors, which regard pixels as positive or negative samples and regress the coordinates of text boxes directly. However, such regression-based methods are conducive to multi-oriented text detection, but not kind to arbitrary shape text detection.

\textbf{Arbitrary Shape Text Detection.}
 Flexible representation of arbitrary shape text instance is of great significance. Regression-based methods for arbitrary shape text are not good enough, thus most arbitrary shape text detectors are based on image segmentation.
Through a set of ordered disks, \cite{long2018textsnake} presented a novel and flexible representation for arbitrary shape text instance, but it requires complicated post-processing steps.
 Similarly, \cite{zhu2018textmountain}, \cite{zhang2019look} utilized the geometric properties of text instances. With accelerated post-processing steps, \cite{zhu2018textmountain} generates text polygons after predicting text scores, text center border probability, and text center direction. \cite{zhang2019look} is to predict text region, text center line and border offsets and furthermore generates text polygons.
All of \cite{long2018textsnake}, \cite{zhu2018textmountain}, \cite{zhang2019look} require additional post-processing steps to generate final text instance by geometric properties of text.
  \cite{wang2019shape} proposed a progressive scale expansion network, and succeeded in detecting arbitrary shape text by predicting masks at multiple scale kernels of text instance, with one clean and efficient post-processing step.
Mask TextSpotter \cite{lyu2018mask} is proposed based on Mask R-CNN. It employed character-level text and word-level text to implement end-to-end text detection and text recognition, but supplementary character-level annotation for character recognition is required. 

Our model is regression-based and pixel-based, which eliminates the limitation of previous pixel-based methods on arbitrary shape text detection and has a simpler pipeline than segmentation-based methods.

\section{Proposed Methodology}
This section begins with an overview of network architecture, 
and then label generation in details will be described in Section 3.2.
The proposed starting-point-independent coordinate regression loss and text instance accuracy loss will be demonstrated in Section 3.3 and 3.4.
In the last, training loss is illustrated.

\subsection{Network Architecture}
As shown in the pipeline of Figure \ref{fig:pp}, a simple and effective text detector for arbitrary shape text with only NMS post-processing step is proposed in this paper.
Our model is fully convolution network.
We use multi-level convolutional feature maps based on SSD with the backbone VGG16. 
As demonstrated in Figure \ref{fig:net}, 7 convolutional feature maps (corresponding to conv4\_3,  fc7 in VGG16 and additional 5 convolution blocks) are fed into category map predictor and coordinates regressor. 
All of the size and stride of each feature map and the range of text-level are indicated in Table \ref{table:fmsize}.
Three stacked convolution layers are applied to category discriminator and coordinate regressor.

\subsection{Label Generation}
As shown in Figure \ref{fig:lg}, the proposed method is based on pixels not anchors, we should obtain target category map and target coordinates map between pixels and arbitrary shape polygons before training.

\begin{table}
\caption{Sizes and strides of feature maps and lower bounds and upper bounds for text level. L0-L6 are corresponding to conv4\_3, conv6, conv7\_2, conv8\_2, conv9\_2, conv10\_2 and conv11\_2 in Figure \ref{fig:net}. Feature map size is abbreviated as fm size and grid size stands for stride of each feature map. Lower and upper are lower and upper bounds of text-level for the corresponding feature map layer.}\smallskip
\centering
\resizebox{1\columnwidth}{!}{
\smallskip\begin{tabular}{|l|ccccccc|}
\hline
Layer & L0 & L1 &L2&L3&L4&L5&L6\\
\hline
fm size & 64 &32  &16 &8&6&4&2\\
grid size & 8 & 16 &32 &64&85&128&256\\
lower& 1.2 &7.2  &14.4 &28.8&38.9&57.6&115.2\\
upper&10.0  &20.0  &35.2 &49.0&85.4&140.8&268.8\\
\hline
\end{tabular}
}
\label{table:fmsize}
\end{table}

\textbf{Classification Label Generation.}
As shown in Figure \ref{fig:lg}, a pixel inside the text will be positive. 
First of all, we need to assign the ground truths to the corresponding feature map layer due to the structure of SSD. Here, we calculate the ratio of polygonal area to polygonal perimeter and regard the ratio as the reference level of text (this manner is available for arbitrary shape text). The lower and upper bounds of text-level for each feature map can be found in Table \ref{table:fmsize}.
After the text-level of each polygon is assigned, only the pixels inside the ground-truths will be regarded as positive samples.

\textbf{Coordinates Map Generation.}
The coordinates map is the target of coordinates regression task.
As shown in Figure \ref{fig:lg}, what we need to regress is the offset between positive pixel and vertex.
 For an arbitrary polygon \textbf{P} (concave or convex), there is a positive pixel \textbf{p} inside \textbf{P} in the $k_{th}$ convolutional layer.
 In this paper, by using the starting-point-independent coordinates regression loss in Section 3.3 instead of conventional regression loss, we do not need to determine the starting point of a polygon or the starting edge of text, but only need to ensure the vertices of polygon \textbf{P} are ordered clockwise.
 Since the number of vertices in the original dataset may not be uniform, we will sample them into a uniform set of \textbf{n} points. In this paper, we select \textbf{n}=4 for multi-oriented text dataset and \textbf{n}=16 for arbitrary shape text dataset.
The sampled coordinate set of  \textbf{P} is as follows:
\begin{equation}
\begin{split}
C_{P} = &\{x_{1}, y_{1}, x_2, y_2,...., x_n, y_n\},
\end{split}
\end{equation}\label{form:loc}
\noindent let the coordinates of \textbf{p} in $k_{th}$ level be $\{x_{p}, y_{p}\}$, 
thus the coordinates will be normalized to
\begin{equation}
\begin{split}
& x^{*}_{i} = \frac{x_{i}-x_{p}}{grid\_size_{k}},  \\
& y^{*}_{i} = \frac{y_{i}-y_{p}}{grid\_size_{k}}, i \in \mathbb{N}, i \in [0, n].
\end{split}
\end{equation}\label{form:reg}

\subsection{Starting-point-independent Coordinates Regression Loss}
In this paper, the starting-point-independent coordinates regression loss is proposed to avoid determination of the starting-point and achieve the direct coordinate regression of arbitrary shape text instances.

\begin{figure}
\centering
\subfigure[]{
\includegraphics[width=0.22\textwidth]{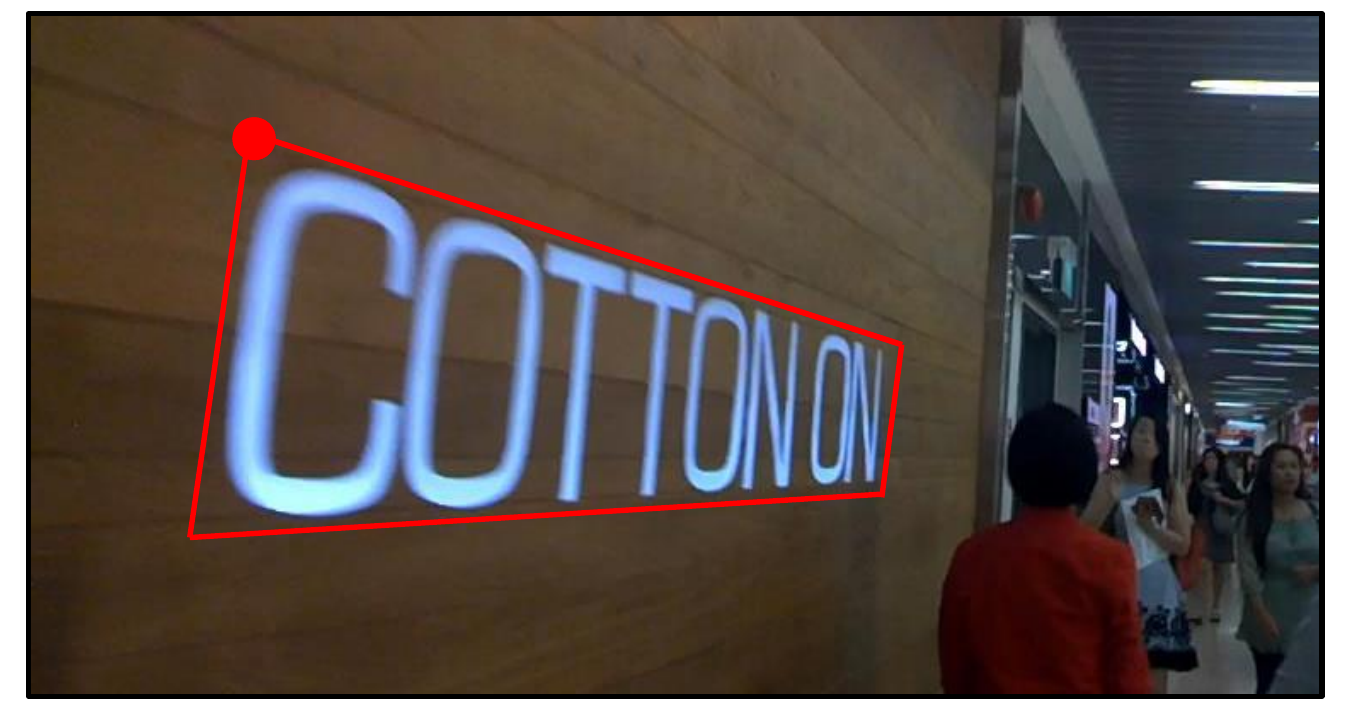}
}
\subfigure[]{
\includegraphics[width=4cm,height=2.07cm]{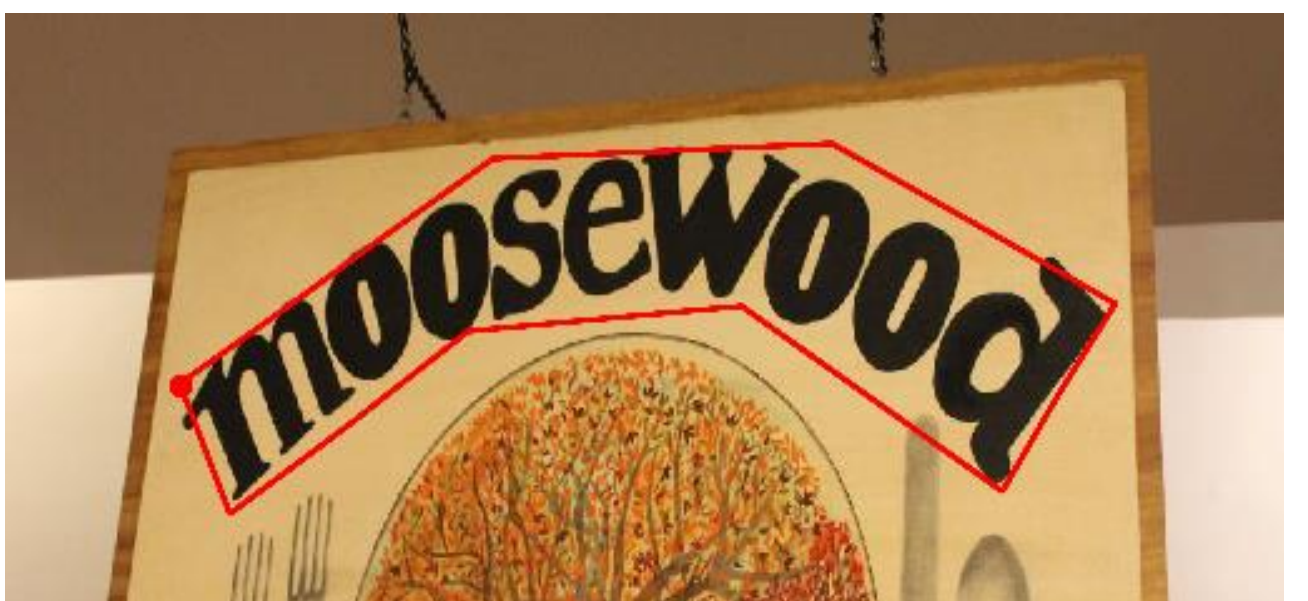}
}
\subfigure[]{
\includegraphics[width=0.22\textwidth]{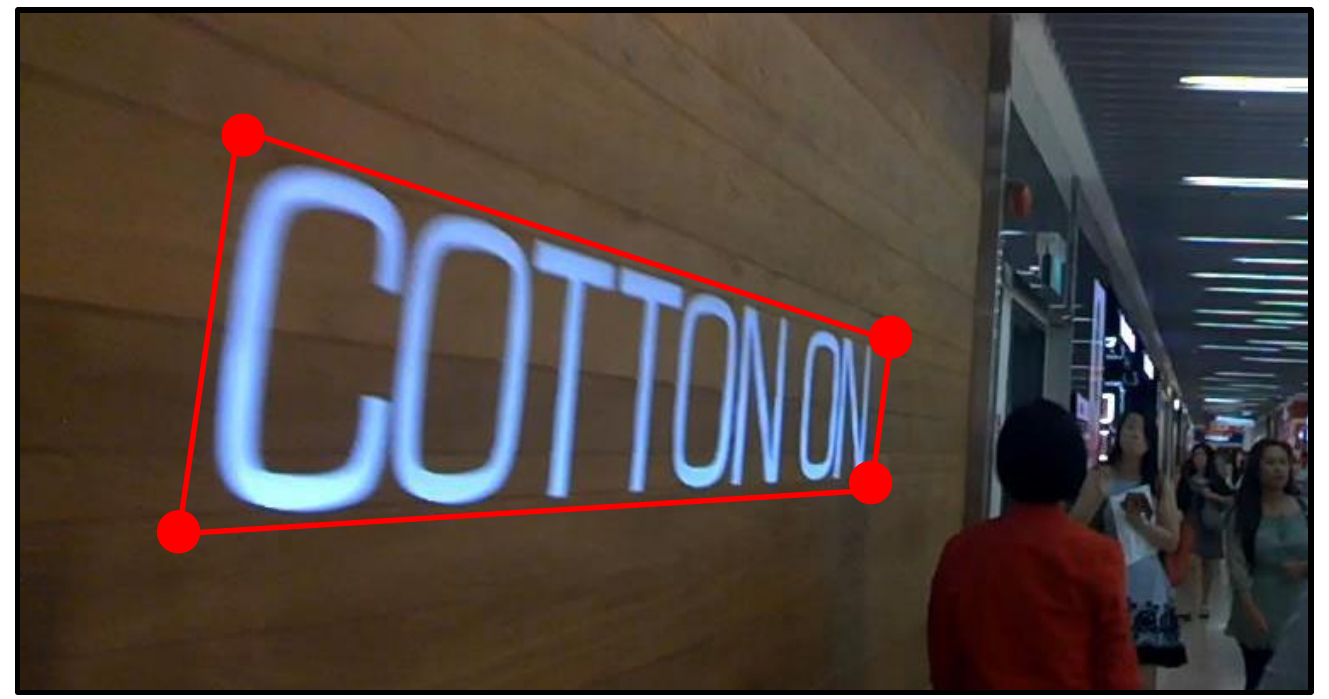}
}
\subfigure[]{
\includegraphics[width=4cm,height=2.12cm]{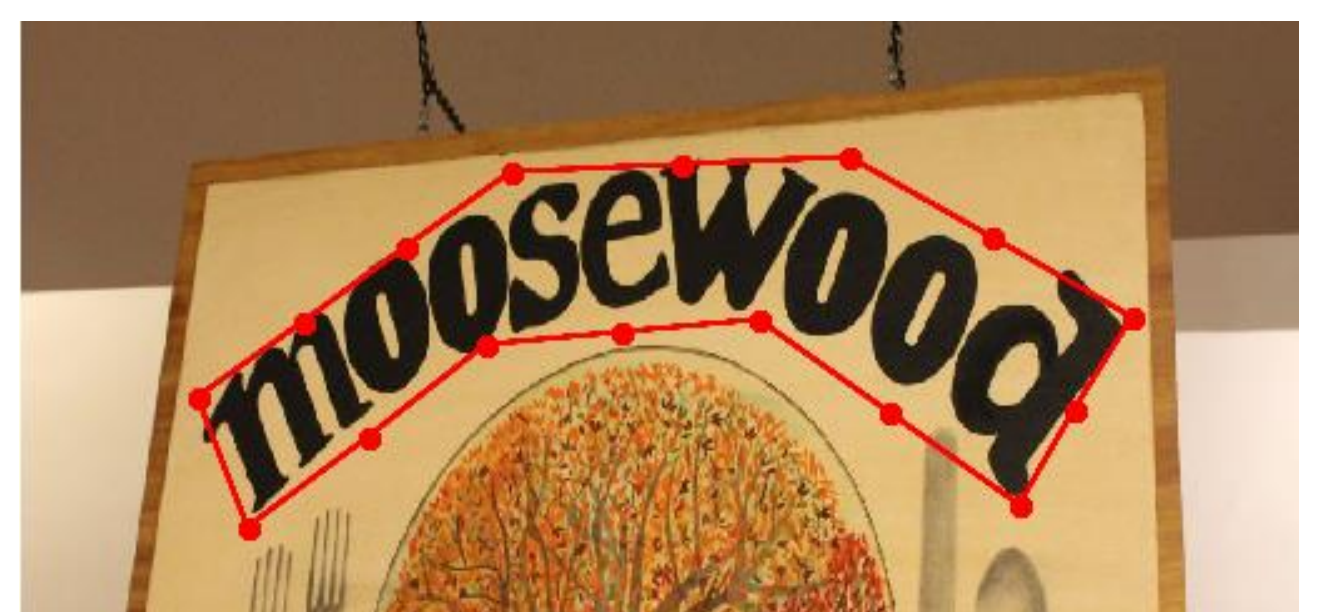}
}
\caption{Starting-point-dependent polygon and starting-point-independent polygon. Red points are the starting point candidates. (a) (b) are in original regression loss, the points need to determine the starting point and be sorted in a clockwise order. (c) (d) are in starting-point-independent coordinate regression loss and the points only need to be sorted in a clockwise order. }\label{fig:ps}
\end{figure}

Smooth L1 loss is widely used for regressing coordinates of text boxes.
From previous work, such as \cite{zhou2017east}, we can see that the vertices of text boxes should be sorted according to a specific pattern which will determine the starting-point in the vertex sequence.
There are some limitations in defining the starting point artificially. In \cite{zhou2017east}, the top left corner vertex will be the starting point directly, as shown in Figure \ref{fig:ps} (a), (b).
As for quadrangles in multi-oriented text detection, the top left corner vertex will be the starting point, which is easily determined. 
Nevertheless, it is a hot thing to define `top-left'/`starting-point' for arbitrary shape polygons, especially concave polygons with large curvature.
That is one of the reasons why some pixel-based methods are unfriendly to arbitrary shape text and some segmentation-based methods
introduce the geometric properties of text instance to represent text instances with arbitrary shapes.

Concerning the above situation, we propose the starting-point-independent coordinates regression loss to make the pixel-based model workable for arbitrary shape text detection. As shown in Figure \ref{fig:ps} (c), (d), the detector can not be affected by the order of vertices and the starting point will be determined adaptively.
The following is the proposed regression loss function for polygon vertices:

\begin{equation}\label{form:loc_loss}
\begin{split}
L_{reg}= \sum_{m \in L_{reg}^{+} } \min_{j\in [0,...,n-1]} \sum_{i=0}^{n-1} smooth_{L_1}(\hat{z}_{i}^{m} - {z^{m*}_{(j+i)\%n}})
\end {split}
\end{equation}

\begin{equation}\label{form:smooth_loss}
smooth_{L_1}(x)=\left\{
    \begin{array}{lr}
    0.5x^2  &\text{ if  } |x| < 1, \\
    |x|-0.5 & \text{otherwise}.
    \end{array}
\right.
\end{equation}
where $\hat{z}^{m}_{i}$ is the $i_{th}$ vertex of the $m_{th}$ predicted polygon, $z^{m*}_{i}$ is the $i_{th}$  vertex of the $m_{th}$ ground truth, $m \in L_{reg}^{+} $ means
$m$ is the element of positive ground truth labels.

\begin{figure*}
\centering
\includegraphics[width=0.29\textwidth,height=2.8cm]{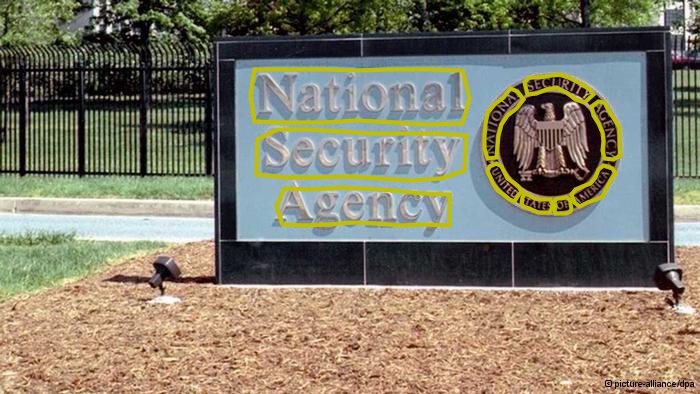}
\includegraphics[width=0.29\textwidth,height=2.8cm]{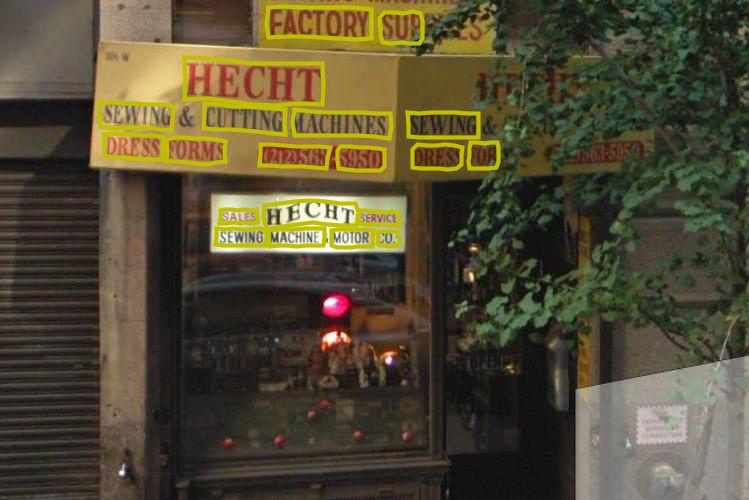}
\includegraphics[width=0.29\textwidth,height=2.8cm]{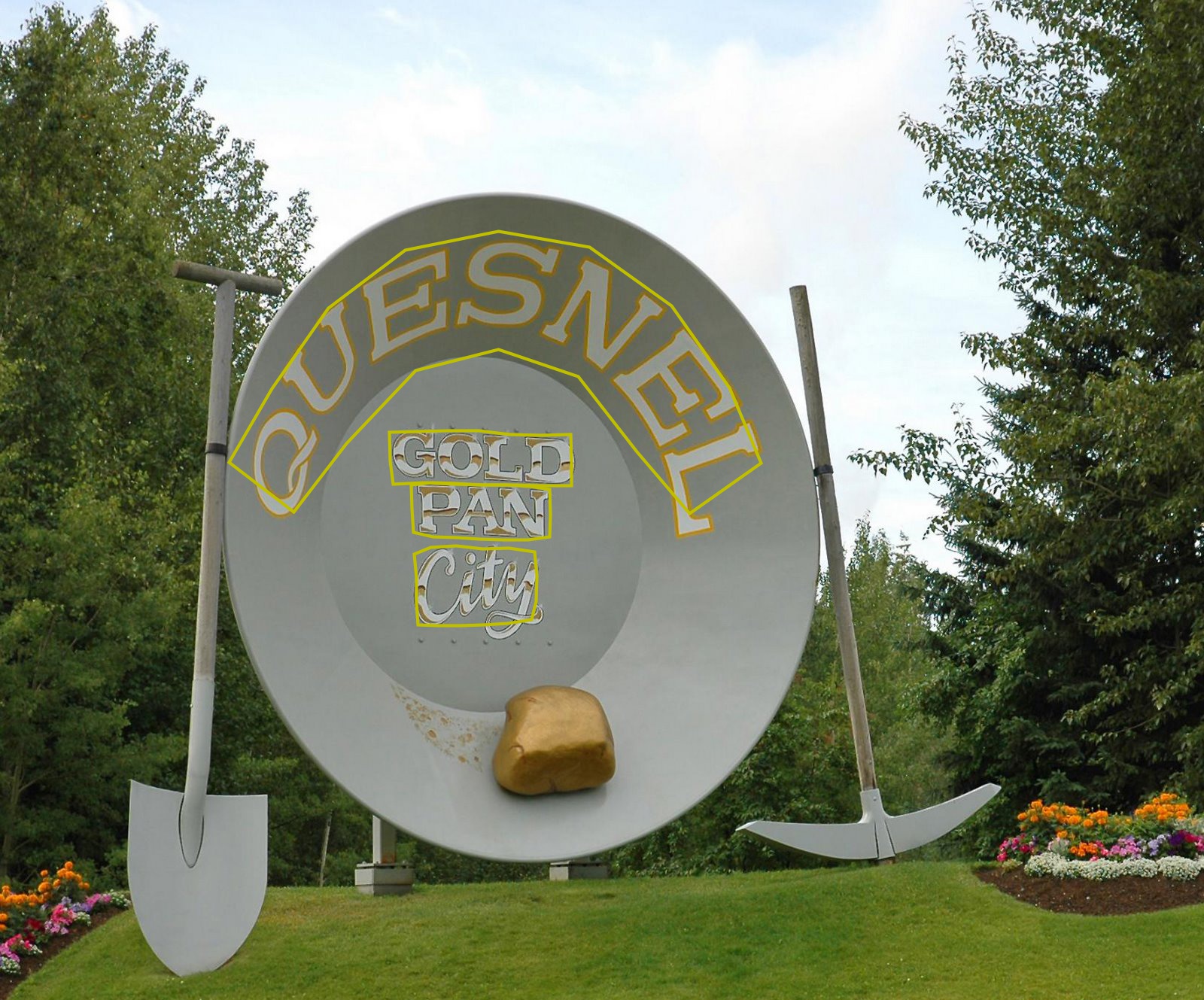}
\includegraphics[width=0.29\textwidth,height=2.8cm]{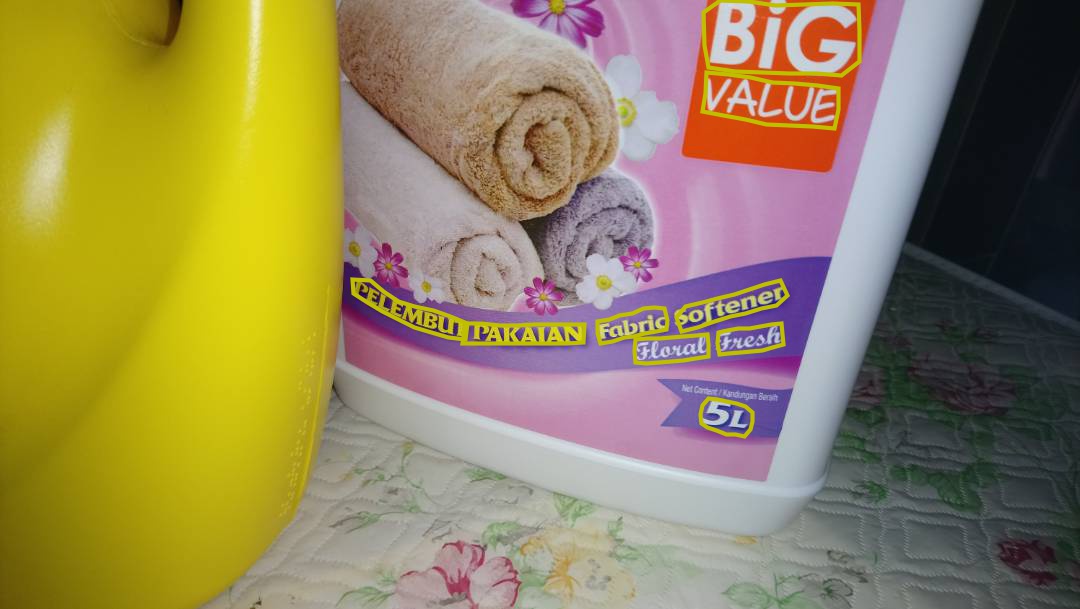}
\includegraphics[width=0.29\textwidth,height=2.8cm]{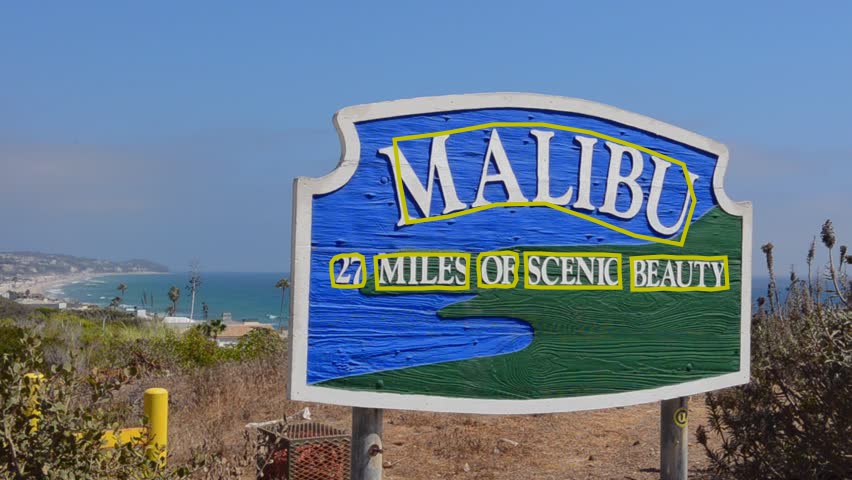}
\includegraphics[width=0.29\textwidth,height=2.8cm]{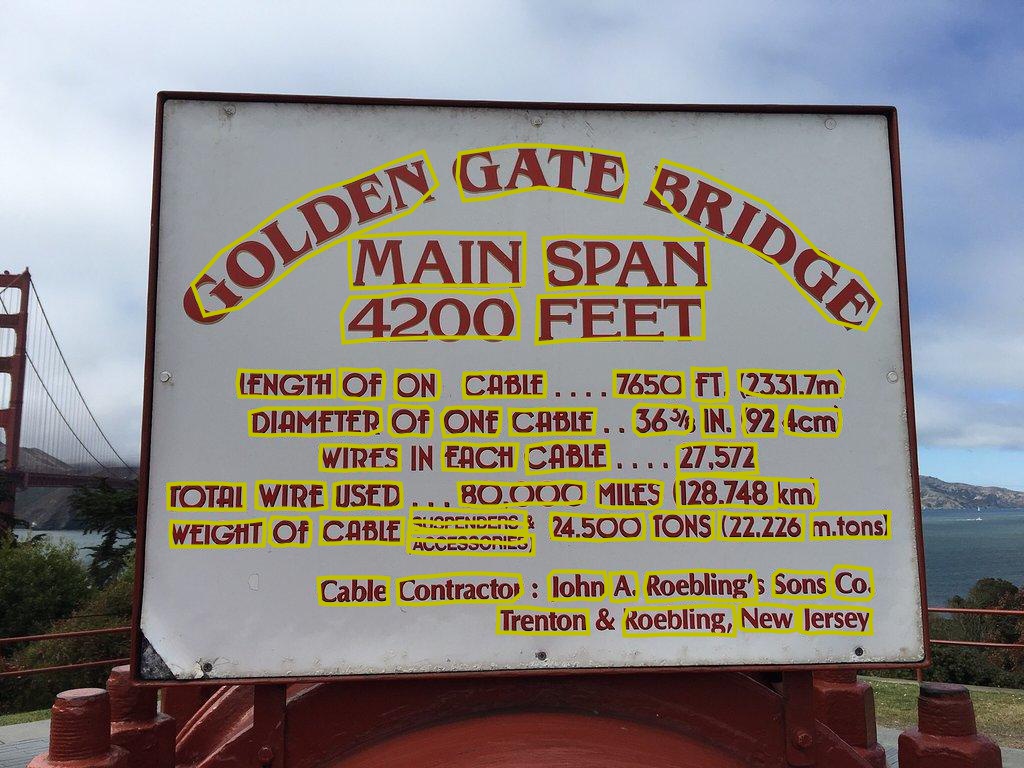}
\caption{Detection examples on Total-Text dataset. The yellow polygons are our detection results. Good detection can be obtained for arbitrary shape text.}\label{fig:ttres}
\end{figure*}

\subsection{Text Instance Accuracy Loss}
To align the predicted polygons with the ground truths and obtain more accurate predicted polygons,
 we introduce the text instance accuracy loss inspired by \cite{ling2019fast}.
The text instance accuracy loss is differential so that it can guide the network to refine the predicted coordinates during the training phase, and does not take up time during inference.

 As shown in Figure \ref{fig:diff}, this loss function is an operation on ground truth polygons and the output coordinates of network. 
 For a positive pixel, there is a ground truth polygon and a  coordinate sequence of predicted polygon.
The rendered mask will generate by the rendering process, which is defined by $M_{P}=R(P)$, where $P$ is the prediction sequence of points sets for a polygon, $R$ is the rendering function, and $M_{P}$ is the mask rendered from $P$.

The size of rendered masks and ground truth masks will be normalized into $64 \times 64$ in this paper. Then,
 L1 loss in formula (\ref{form: Diff_loss}) is employed in calculating the loss between rendered mask $M_{P}$ and the ground truth mask $M_{gt}$.
\begin{equation}\label{form: Diff_loss}
L_{acc} = \sum \Vert M_{P} - M_{gt}\Vert_{1}
 \end{equation}

 From the above statements, $L_{acc}$ is exactly pixel-wise accuracy and related to IoU between predicted polygon and ground truth polygon. The smaller $L_{acc}$ is, the bigger overlap area is.
 The text instance accuracy loss is to refine coordinates of predicted polygons under the guidance of IoU, however the starting-point-independent loss optimize coordinates directly.
 The experimental results in Table \ref{tab:abl} prove the effectiveness of text instance accuracy loss. Based on the batch size of 16 in this paper, random 256 candidates (whose IoU are larger than 0.5) are selected for loss optimization in each iteration.

\subsection{Training Loss}

 The loss function in this paper can be formulated as
\begin{equation}\label{form:loss}
\begin{split}
 Loss =  &  \lambda_{cls} \cdot L_{cls} +\lambda_{reg}\cdot L_{reg} + \lambda_{acc} \cdot L_{acc},
 \end{split}
 \end{equation}
where $L_{cls}$ is the loss of pixel-based classification task, $L_{reg}$ presents starting-point-independent coordinates regression loss, 
 and $L_{acc}$ refers to text instance accuracy loss of more accurate polygon prediction task. $L_{reg}$ and $L_{acc}$ are described in Section 3.3 and Section 3.4. $ \lambda_{cls} $, $ \lambda_{reg} $,  $\lambda_{acc}$ are the corresponding loss-balancing parameters.

The classification loss is to optimize the pixel category and the pixel inside text would be positive.
As positive pixels are far less than negative pixels, Focal loss \cite{lin2017focal} is selected in this paper to alleviate the problem of label imbalance in SSD.
Thus, we formulate $L_{cls}$ as
\begin{equation}\label{form:cls_loss}
 \begin{split}
L_{cls} = &-\sum (y^{*}_{i}\cdot \alpha \cdot (1-\hat{y_{i}}) ^\gamma \cdot log(\hat{y_i}) \\
     &+ (1-y^{*}_{i})\cdot (1-\alpha)\cdot (\hat{y_{i}}) ^\gamma \cdot log(1-\hat{y_i})),
\end{split}
\end{equation}
 where $y_{i}^{*}\in \{0, 1\}$ means classification label for the ${i_{th}}$ pixel, $\hat{y_{i}}$ is predicted class confidence. $\alpha$ and $\gamma$ are parameters of focal loss. In our experiments, $\alpha=0.25$, $\gamma=2$.

%


\subsection{Implementation Details}
The backbone of our network is initialized by the pre-trained model on ImageNet, and the rest convolutional layers are initialized by Xavier uniform.
The network can be end-to-end trainable by the losses in Section 3.5. The weights of losses could be 
 $[\lambda_{cls}, \lambda_{reg}, \lambda_{acc}]=[40, 1, 0.01/1]$,
 and $\lambda_{acc}$ is initialized with 0.01 and it is changed into 1 after 60k iterations. The Adam optimizer with learning rate 5e-5, weight decay 1e-6 is to optimize the model.
 Random resizing, random crop and random rotation such data augmentation techniques are necessary to enrich training samples. All these samples are $512 \times 512$  regions cropped from scaled original images, and these samples maintain the original aspect ratio.
With batch size set at 16 in the training phase, all the experiments are implemented in Pytorch by using two NVIDIA GTX 1080Ti GPUs.
In the testing phase,
pixels with classification confidence above 0.7 will be treated as
text. Standard non-maximum suppression with overlap ratio 0.3 is available in this paper.

\section{Experiments}
In this section, we conduct an extensive evaluation of our model on Total-Text dataset, SCUT-CTW1500 dataset and ICDAR 2015 Incidental Scene dataset. 
Moreover, we analyze the experimental results and present a comparison with other state-of-the-art methods.

\subsection{Datasets}
Our experiments are mainly based on English text datasets. The benchmarks involve arbitrary shape text of word-level, arbitrary shape text of textline-level and multi-oriented text of word-level. 

\textbf{Total-Text.} Total-Text \cite{ch2017total} has 1,555 images (1,255 for training, 300 for testing), and they are from business-related, tourist spots, formal information, and club logos and others.
The dataset contains horizontal, multi-oriented, and curved text in word-level. Polygon-shaped ground truths with the inconsistent number of vertices are provided for text instances.

\textbf{SCUT-CTW1500.} There are 1,500 images (1,000 images for training, and 500 images for testing), that come from the Internet, Google's open-Image and mobile phone cameras. Each image contains at least one curved text.
The text-line instances are mainly arbitrary shape texts, and it also contains horizontal or multi-oriented text-lines.
The dataset provides the same number of 14 points for each polygon.

\textbf{ICDAR2015 Incidental Scene Text.} ICDAR 2015 Incidental Scene Text \cite{karatzas2015icdar} consists of 1,500 images (1,000 images for training and 500 images for testing), which are randomly captured
by Google glasses. The text instances in word-level are multi-oriented and scattered randomly in the images, and each instance has an annotation of 8 coordinates of 4 vertices in a clockwise order.

\begin{table}
\caption{Comparisons with other methods on Total-Text. * indicates results from \cite{long2018textsnake}. }
\resizebox{0.46\textwidth}{!}{
\begin{tabular}{l|c|c|c}
\hline
Method & Precision & Recall & F-measure \\
\hline\hline
SegLink$^*$\cite{shi2017detecting} &30.3$^*$ &23.8$^*$& 26.7$^*$\\
EAST$^*$\cite{zhou2017east}& 50.0$^*$&36.2$^*$&42.0$^*$\\
TextBoxes$^*$  \cite{liao2017textboxes}&62.1$^*$  &45.5$^*$ &52.5$^*$ \\
Mask TextSpotter\cite{lyu2018mask} &69.0 &55.0 &61.3 \\
PSENet-1s \cite{wang2019shape}&81.8&75.1&78.3\\ 
TextSnake\cite{long2018textsnake}&82.7 &74.5&78.4 \\
LOMO \cite{zhang2019look}&88.6 &75.7&81.6\\
CRAFT\cite{baek2019character}&87.6 &79.9&83.6\\ 
\hline\hline
s-reg &86.6 &81.4&83.9\\
s-reg+acc &\textbf{87.7} &\textbf{82.0}&\textbf{84.8}\\
\hline
\end{tabular}}
\label{tab:total}

\end{table}

\begin{table}
\caption{Comparisons with other methods on SCUT-CTW1500. * indicates the results from \cite{yuliang2017detecting}. }
\resizebox{0.46\textwidth}{!}{
\begin{tabular}{l|c|c|c}
\hline
Method &Precision & Recall & F-measure \\
\hline\hline
CTPN$^*$ \cite{tian2016detecting}&60.4$^*$& 53.8$^*$ &56.9$^*$\\
SegLink$^*$\cite{shi2017detecting}&42.3$^*$& 40.0$^*$ &40.8$^*$\\
EAST$^*$\cite{zhou2017east}& 78.7$^*$ &49.1$^*$& 60.4$^*$\\
CTD+TLOC\cite{yuliang2017detecting}&77.4 &69.8 &73.4\\
TextSnake\cite{long2018textsnake}&67.9&85.3&75.6 \\
LOMO \cite{zhang2019look}&89.2 &69.6&78.4\\
PSENet-1s \cite{wang2019shape}&80.6&75.6&78.0\\ 
CRAFT\cite{baek2019character}&\textbf{87.6} &\textbf{79.9}&\textbf{83.6}\\ 
\hline\hline
s-reg &82.0&71.0&76.0\\
s-reg+acc &84.8&71.7&77.7\\
\hline
\end{tabular}}
\label{tab:ctw}
\end{table}

\begin{table}
\caption{Comparisons with other methods on ICDAR2015.}
\resizebox{0.46\textwidth}{!}{
\begin{tabular}{l|c|c|c}
\hline
Method& Precision & Recall & F-measure \\
\hline\hline
SSTD\cite{he2017single} &80.2 &73.9&76.9 \\
EAST\cite{zhou2017east} &83.3&78.3&80.7\\
DDR\cite{he2017deep}&82.0 &80.0&81.0\\
PixelLink\cite{deng2018pixellink}& 85.5&82.0&83.7\\
PSENet-1s \cite{wang2019shape}&81.5&79.7&80.6\\ 
TextSnake\cite{long2018textsnake} &84.9&80.4&82.6\\
Mask TextSpotter\cite{lyu2018mask} &\textbf{91.6}&81.0&86.0\\
CRAFT\cite{baek2019character}&89.8&84.3&86.9\\ 
LOMO\cite{zhang2019look}&87.8 &\textbf{87.6}&\textbf{87.7}\\
\hline\hline
s-reg&86.3 &84.3&85.2\\
s-reg+acc & 88.2&84.8&86.5\\
\hline
\end{tabular}}
\label{tab:ICDAR15}

\end{table}

\subsection{Experimental Results}
%

%

Here we present the experimental results on public benchmarks. The experimental results demonstrate that our text detector is simple and effective.

\textbf{Total-Text.}
As shown in Table \ref{tab:total}, our method outperforms segmentation-based methods, such as \cite{long2018textsnake}, \cite{wang2019shape}, \cite{baek2019character}. The proposed method achieves 84.8\% of F-measure, 86.3\% of precision and 83.4\% of recall on Total-Text dataset, and reaches state-of-the-art performance, to our best knowledge. It demonstrates that the proposed method utilizes simple and intuitive tasks to train the model and obtains the effective detector on arbitrary shape text dataset.
Some detection samples on Total-Text are indicated in Figure \ref{fig:ttres}.

\begin{figure*}
\centering
\includegraphics[width=0.29\textwidth,height=2.8cm]{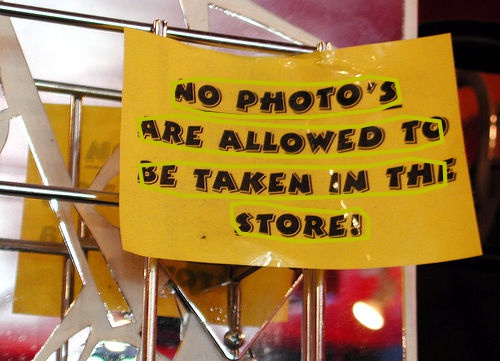}
\includegraphics[width=0.29\textwidth,height=2.8cm]{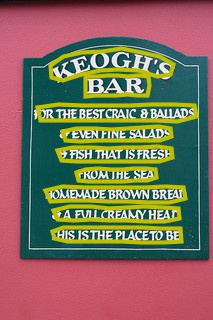}
\includegraphics[width=0.29\textwidth,height=2.8cm]{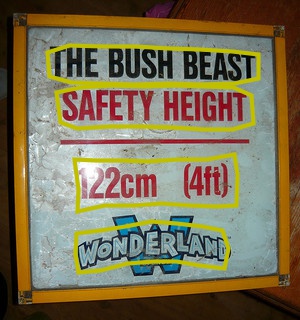}
\includegraphics[width=0.29\textwidth,height=2.8cm]{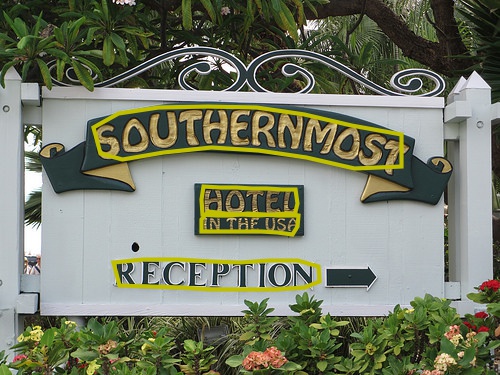}
\includegraphics[width=0.29\textwidth,height=2.8cm]{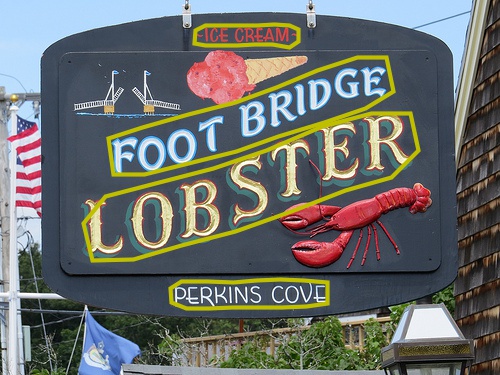}
\includegraphics[width=0.29\textwidth,height=2.8cm]{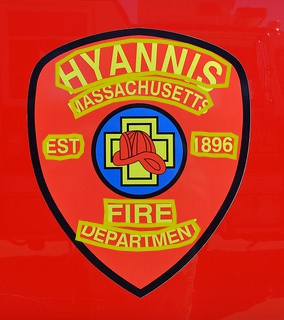}
\caption{Detection examples on SCUT-CTW1500 dataset. The yellow polygons are our detection results.}\label{fig:ctwres}
\end{figure*}

\textbf{SCUT-CTW1500.}
As shown in Table \ref{tab:ctw}, the comparisons of performance on SCUT-CTW1500 are demonstrated, and we obtain 77.7\% of F-measure. 
 As a direct regression method, our method performs better than \cite{zhou2017east}.
 Beyond that, our performance on SCUT-CTW1500 is better than \cite{yuliang2017detecting}, and \cite{long2018textsnake},
but is slightly lower than some methods \cite{zhang2019look}, \cite{wang2019shape}, \cite{baek2019character}. Some detection instances of our model are shown in Figure \ref{fig:ctwres}.

\begin{figure*}
\centering
\includegraphics[width=0.29\textwidth]{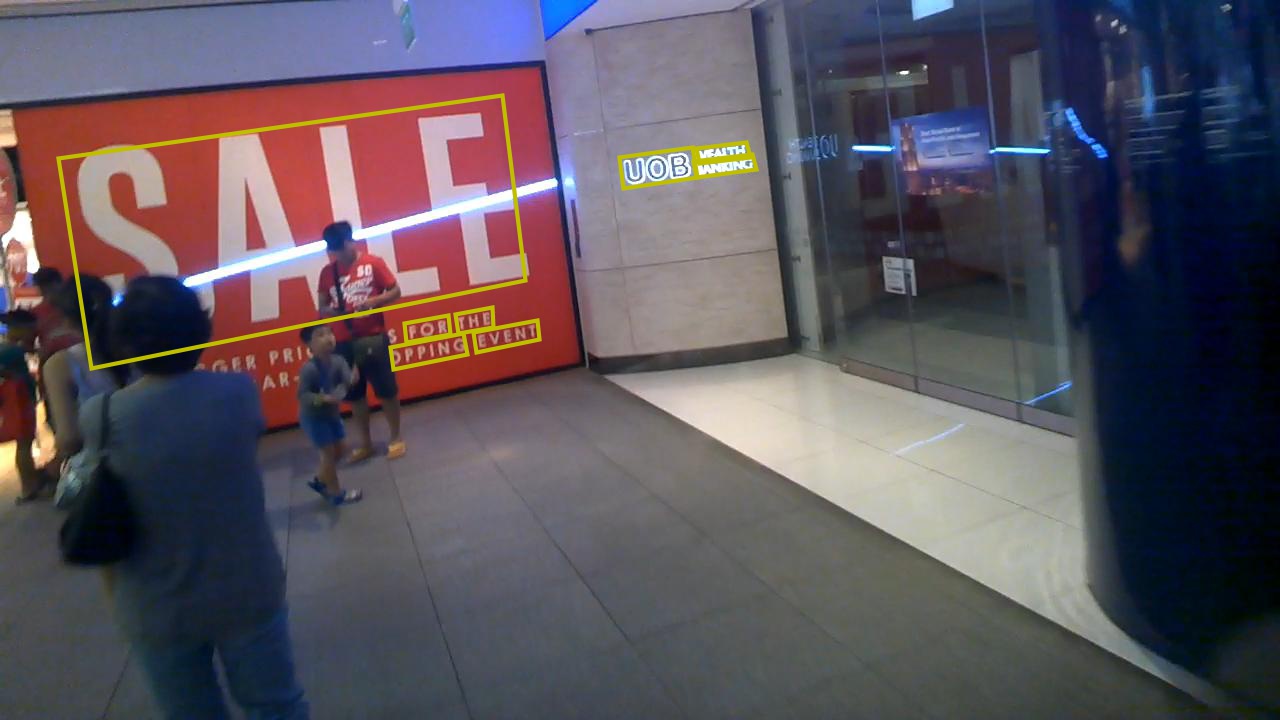}
\includegraphics[width=0.29\textwidth]{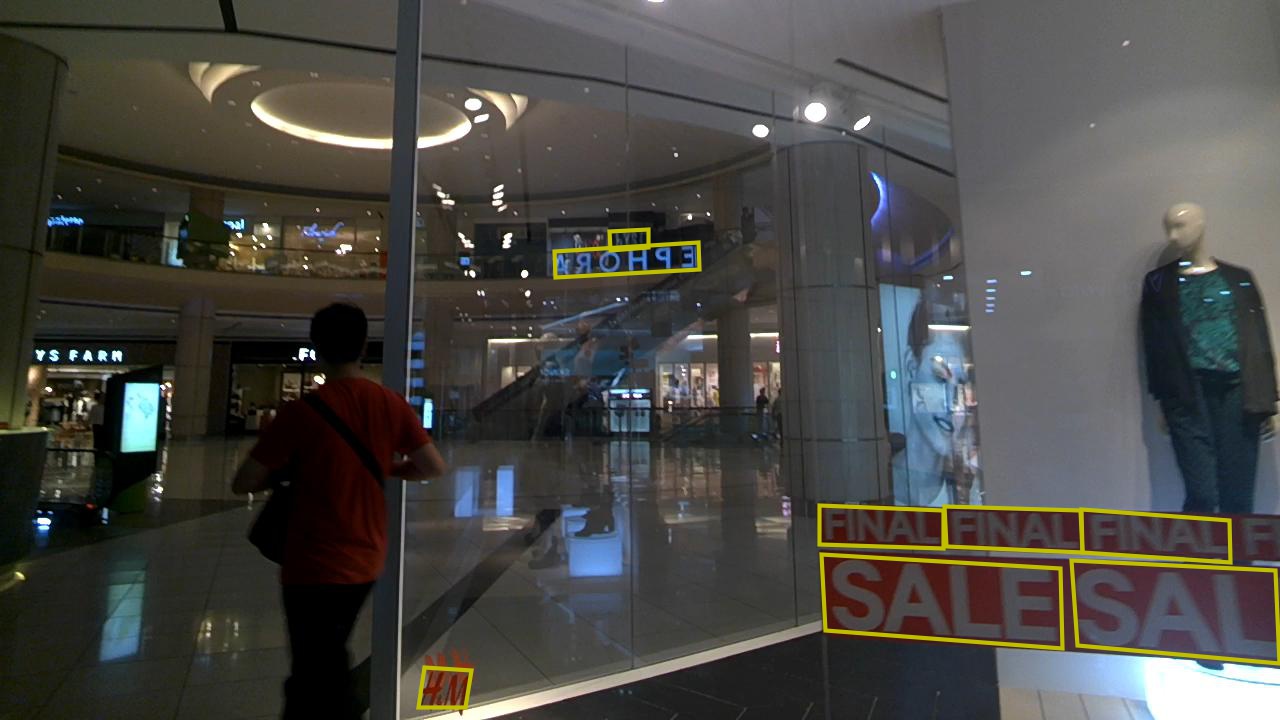}
\includegraphics[width=0.29\textwidth]{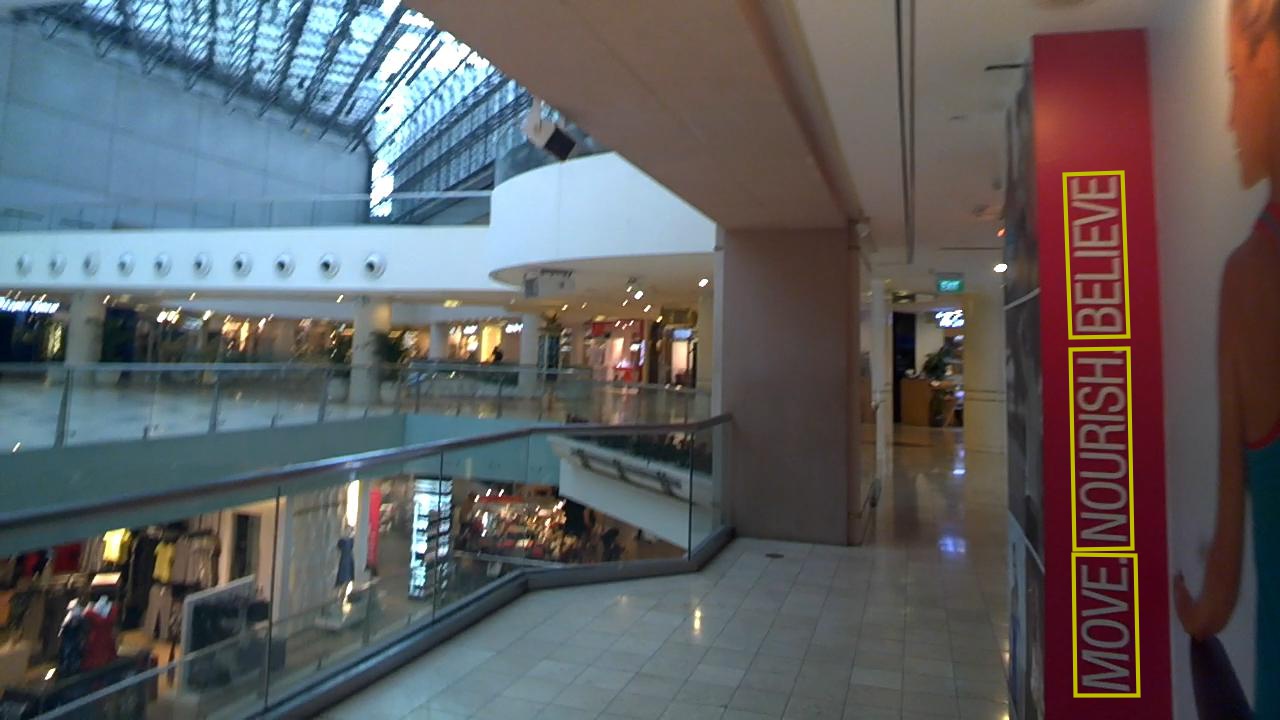}
\includegraphics[width=0.29\textwidth]{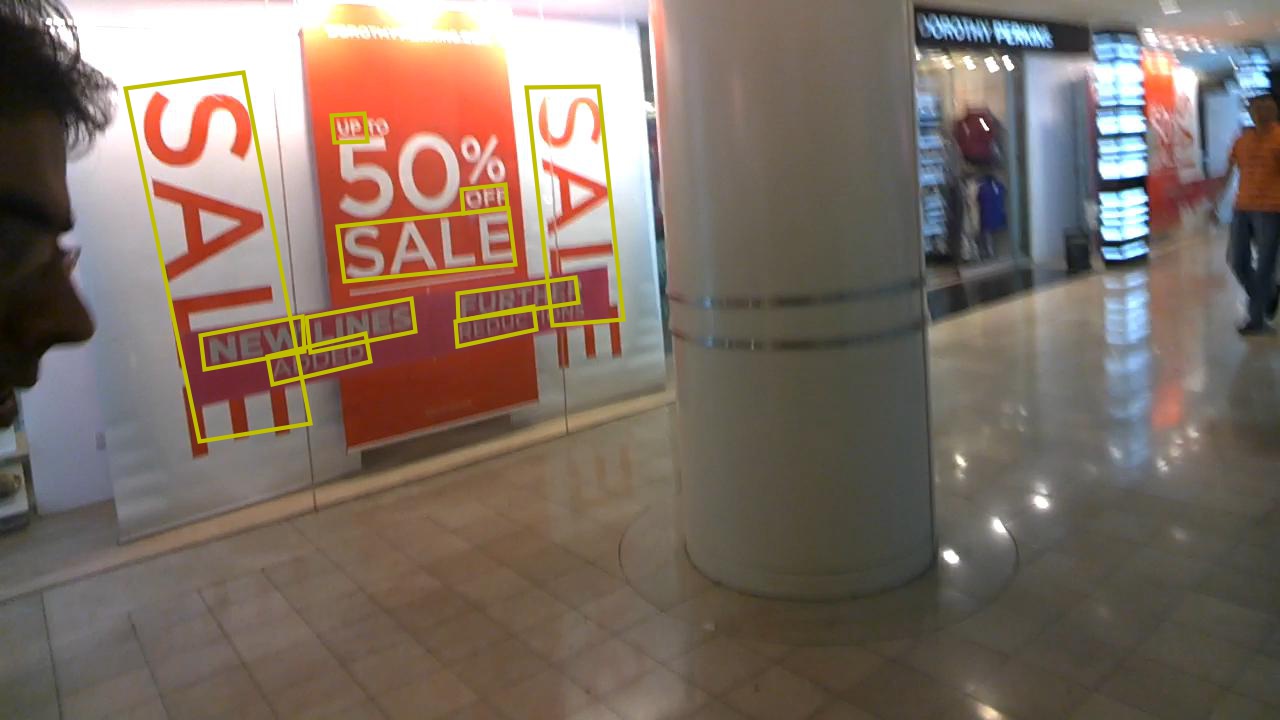}
\includegraphics[width=0.29\textwidth]{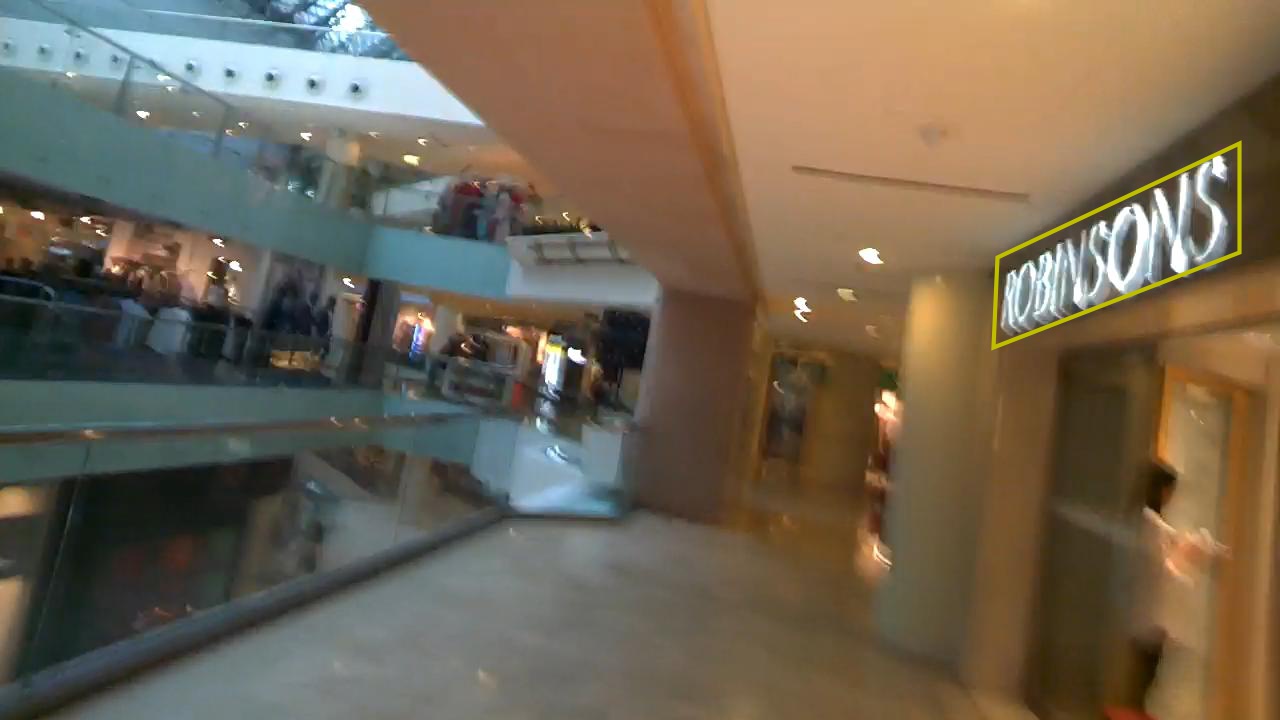}
\includegraphics[width=0.29\textwidth]{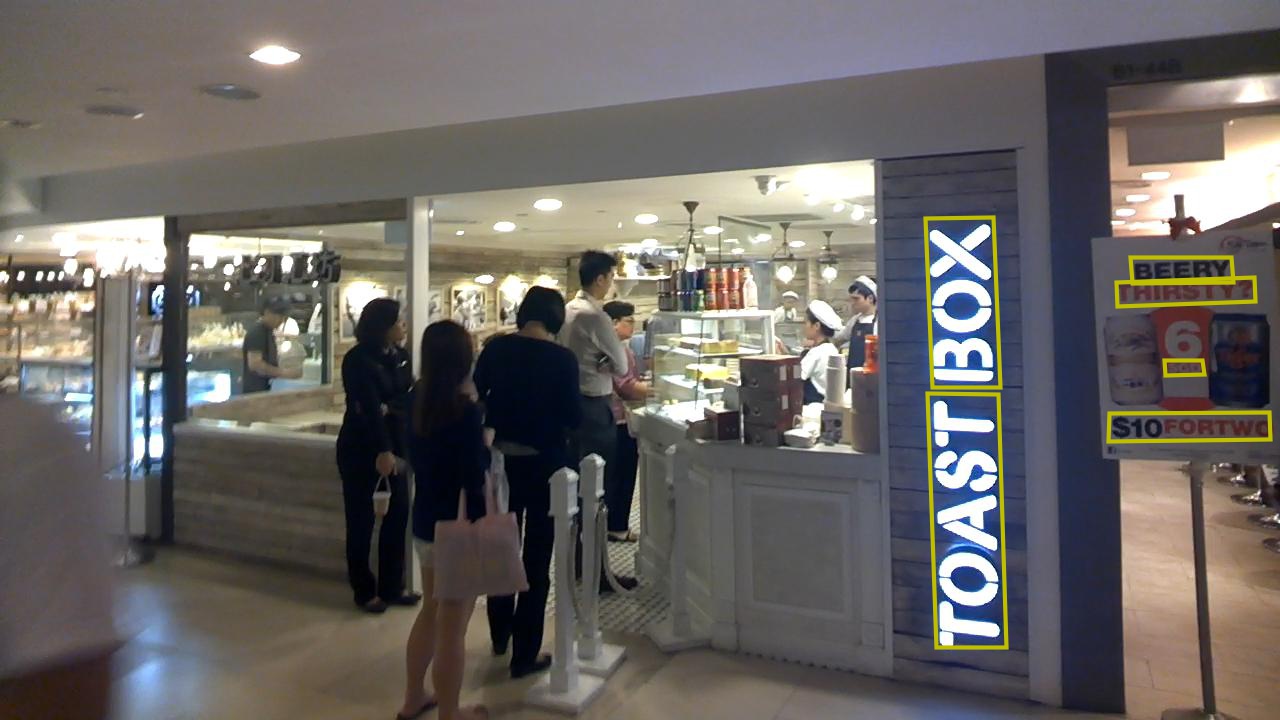}
\caption{Detection examples on ICDAR 2015 Incidental Scene Text dataset. The yellow polygons are our detection results. From the figure,  our method can perform well on horizontal, vertical and slanted text.}\label{fig:15res}
\end{figure*}

\textbf{ICDAR2015 Incidental Scene Text.} The performance of our method on ICDAR 2015 Incidental Scene Text is illustrated in Table \ref{tab:ICDAR15}, we obtain 86.5\% of F-measure and reach a good performance. As for multi-oriented text detection, our method is pixel-based. Our method is free of designing anchors and performs better than \cite{ma2018arbitrary}, \cite{jiang2017r2cnn} 
which are anchor-based.
We achieve a better performance than \cite{zhou2017east}, \cite{he2017deep} which are pixel-based as well. Furthermore, we employ more easier representation of text and reach a better performance than \cite{long2018textsnake}, \cite{lyu2018mask}. Our performance on ICDAR2015 is slightly lower than \cite{baek2019character} and \cite{zhang2019look}.
In Figure \ref{fig:15res}, there shows some detection examples.

\subsection{Ablation Study}
Here we will evaluate the effectiveness of  the proposed loss functions on ICDAR2015 Incidental Scene Text and total text through ablation study.

\textbf{Starting-point-independent Coordinates Regression Loss.}
  In this part, we compare the conventional regression loss with the starting-point-independent coordinates regression loss on ICDAR 2015 Incidental Scene Text dataset.
  As shown in Table \ref{tab:abl}, the starting-point-independent coordinates regression loss function obtains 0.5\% of F-measure of improvement on multi-oriented text detection. The improvement means that there are some imperfections in determining the starting point, and proves the effectiveness of the proposed regression loss.

  Due to the hard definition of starting-point in arbitrary shape polygon, we only evaluate the effectiveness of starting-point-independent coordinates regression loss function on Total-Text dataset.
As illustrated in Table \ref{tab:abl}, the proposed regression loss makes our model available for arbitrary shape text with a simple pipeline and reaches 84.8\% of F-measure on arbitrary shape text detection which achieves the state-of-the-art performance.

\textbf{Text Instance Accuracy Loss.}
 As can be seen from Table \ref{tab:abl}, the text instance accuracy loss can assist the model have 1.3\% improvement of F-measure on ICDAR 2015 Incidental Scene Text dataset and 0.9\% improvement on Total-Text dataset.
In addition, we calculate the mean IoU of correctly detected text boxes. In Table  \ref{tab:abl}, the mean IoU becomes larger when the model is trained with text instance accuracy loss.
The above improvements demonstrate that text instance accuracy loss is workable in predicting text polygons with larger IoU, and is helpful to further improve F-measure.

\begin{table}
\centering
\caption{Ablation study on public benchmarks, \textbf{reg} means \textbf{conventional regression loss}, \textbf{s-reg} means \textbf{starting-point-independent coordinates regression loss}, 
and \textbf{acc} means \textbf{text instance accuracy loss}. P, R, F mean precision, recall, and F-measure respectively. mIoU is the mean IoU of correctly detected text boxes.}
\resizebox{0.47\textwidth}{!}{
\begin{tabular}{c|cccc|cccc}
\hline
\multirow{2}{*}{Loss}& \multicolumn{4}{|c|}{Total-Text}& \multicolumn{4}{|c}{ICDAR2015}\\
& P& R &F&mIoU& P &R &F&mIoU \\
\hline\hline
reg&-&-&-&-&87.0 &82.5&84.7&0.767\\
s-reg&86.6&81.4&83.9&0.768&86.3 &84.3&85.2&0.767\\
s-reg+acc&\textbf{87.7}&\textbf{82.0}&\textbf{84.8}&\textbf{0.783}&\textbf{88.2}&\textbf{84.8}&\textbf{86.5} &\textbf{0.775} \\
\hline
\end{tabular}}\label{tab:abl}
\end{table}
%

 \subsection{Speed}
We test the time consumption on a GeForce GTX 1080Ti graphic card and Intel(R) Xeon(R) E5-2620 v3 @2.40GHz CPU.
During the testing phase, we use the original images ($1280\times 720$) of ICDAR2015 Incidental Scene Text dataset.
The time consumption is computed by the average running time for images in the test dataset.
Text instance accuracy loss doesn't participate in inference, hence 
models trained with or without text instance accuracy loss have the same time consumption.
As the comparisons of time consumption shown in Table \ref{tab:sp}, our model runs at 7.72 FPS on ICDAR 2015 Incidental Scene Text.

%
%

\begin{table}
\centering
\caption{Comparisons of time consumption on ICDAR 2015 Incidental Scene Text during inference. Num. means the number vertices of predicted polygon, Res. is the resolution of the input image. }
\resizebox{0.42\textwidth}{!}{
\begin{tabular}{c|c|c|c}
\hline
 Method &GPU & Res.  & FPS \\
\hline\hline
TextSnake\cite{long2018textsnake}&Titan X&720p&1.1 \\
PSENet-1s \cite{wang2019shape}&1080Ti&720p&1.6\\ 
EAST$-vgg16$\cite{zhou2017east}& Titan X&720p&6.5\\
\hline\hline
Our & 1080Ti &720p&7.72\\
\hline
\end{tabular}}\label{tab:sp}
\end{table}

 \subsection{Comparison with the state-of-the-art methods}
 Here we have some comparison with some recent methods on text detection. 
 The pipelines are demonstrated in Figure \ref{fig:pp}.
Our method is pixel-based, and we propose the starting-point-independent coordinates regression loss and text instance accuracy loss.
In general, our method can handle with arbitrary shape text detection better than previous pixel-based methods, and than segmentation-based methods, our method has simpler post-precessing steps and a simpler pipeline.

 \textbf{EAST.} Similarly, EAST \cite{zhou2017east} is a pixel-based method and has a consistent pipeline with ours as shown in Figure \ref{fig:pp}. 
 Compared with EAST and other pixel-based method \cite{he2017deep}, our method achieves arbitrary shape text detection well by using the starting-point-independent coordinates regression loss.

 \textbf{TextSnake.} \cite{long2018textsnake} presented a novel, flexible representation of arbitrary shape text instance, which differs from ours. We directly regress the coordinates of text instances which conforms to training annotations, however TextSnake utilized geometric properties to represent text instances. 
  Different from TextSnake, our method is regression-based, not segmentation-based.

 \textbf{LOMO.} LOMO \cite{zhang2019look} is based on regression and segmentation. Before final text polygons are generated, LOMO will produce text rectangular boxes and text geometric property maps.
 Our method is regression-based and predicts text polygons directly without intermediates.

\textbf{CRAFT.} \cite{baek2019character} is a bottom-up method, which generates prediction of character-level. 
However, our method doesn't need to utilize Synthetic text dataset and performs text detection in a top-down manner. Furthermore, our pipeline is simpler.

\textbf{PSENet.} \cite{wang2019shape} proposed a segmentation-based method PSENet which predicts text masks in different scales.
However, our method is regression-based and have a simpler post-processing step.

\section{Conclusion}
Different from previous methods, we propose a simple and effective method, which does not need additional character annotation and post-processing steps (except NMS), to achieve arbitrary shape text detection, and the proposed method reaches state-of-the-art performance.
 In this paper, the starting-point-independent coordinates regression loss is proposed to avoid the determination of starting point and regress the coordinates of arbitrary shape text directly.
 At the same time, our method makes up the gap in arbitrary shape text detection for pixel-based methods.
 Dedicated to predicting more accurate text polygons, the text instance accuracy loss is introduced to further improve the performance.
 Experimental results prove the effectiveness of our method.

{\small
\bibliographystyle{ieee}
\bibliography{egbib}
}

\end{document}